\documentclass{article}

\PassOptionsToPackage{round}{natbib}
\usepackage[preprint]{neurips_2025}

\usepackage[utf8]{inputenc}
\usepackage[T1]{fontenc}
\usepackage[hidelinks]{hyperref}
\usepackage{url}
\usepackage{booktabs}
\usepackage{amsfonts}
\usepackage{nicefrac}
\usepackage{microtype}
\usepackage{xcolor}
\usepackage{graphicx}

\title{Self-Interpretability: LLMs Can Describe Complex Internal Processes that Drive Their Decisions}
 
\author{%
  Dillon Plunkett \\
  Northeastern University \\
  \texttt{plunkett@g.harvard.edu} \\
  \And
  Adam Morris \\
  Princeton University \\
  \texttt{thatadammorris@gmail.com} \\
  \AND
  Keerthi Reddy \\
  Independent Researcher \\
  \And
  Jorge Morales \\
  Northeastern University \\
}

\begin{document}

\raggedbottom

\maketitle

\begin{abstract}
We have only limited understanding of how and why large language models (LLMs) respond in the ways that they do. Their neural networks have proven challenging to interpret, and we are only beginning to tease out the function of individual neurons and circuits within them. However, another path to understanding these systems is to investigate and develop their capacity to explain their own functioning. Here, we show that i) LLMs can accurately describe quantitative features of their own internal processes during certain kinds of decision-making and ii) that it is possible to improve these capabilities through training. To do so, we fine-tuned GPT-4o and GPT-4o-mini to make decisions in a wide variety of complex contexts (e.g., choosing between condos, loans, vacations, etc.) according to randomly-generated, quantitative preferences about how to weigh different attributes (e.g., the relative importance of natural light versus quiet surroundings for condos). We demonstrate that the LLMs can accurately report these preferences (i.e., the weights that they learned to give to different attributes during decision-making). Next, we demonstrate that these LLMs can be fine-tuned to explain their decision-making even more accurately. Finally, we demonstrate that this training generalizes: It improves the ability of the models to accurately explain how they make other complex decisions, not just decisions they have been fine-tuned to make. This work is a step towards training LLMs to accurately and broadly report on their own internal processes---a possibility that would yield substantial benefits for interpretability, control, and safety.
\end{abstract}

\section{Introduction}

A key challenge in studying large language models (LLMs) is understanding why they do what they do. As with all deep neural networks, the internal representations and operations that drive their behavior are, by default, opaque to human eyes. This is unfortunate. For almost all issues that one might consider most important or concerning about LLMs, better understanding how these systems work would be extraordinarily helpful. Doing so would enable us to better control them \citep{nanda_longlist_2022,bereska_mechanistic_2024}, prevent bias in their behavior \citep{gilpin_explaining_2018,gallegos_bias_2024}, and make informed decisions about when to trust their output or decisions \citep{deeks_judicial_2019}.

Attacks on this problem have generally taken one of two approaches \citep{bereska_mechanistic_2024,danilevsky_survey_2020}. The first approach is to use ``black box'' methods \citep{casper_black-box_2024} that try to understand deep neural networks based on observations of their outputs in response to different inputs---much like a cognitive scientist running behavioral experiments on humans. The second approach is to use ``mechanistic interpretability'' methods \citep{olah_zoom_2020,rai_practical_2024} that crack open the black box and try to reverse-engineer the function of, e.g., individual artificial neurons within---much like a cognitive neuroscientist performing single-unit recordings in human brains.

However, in the case of LLMs, there is a third approach that we could use, the method that people most commonly use to discover the thoughts and motivations of other humans: just asking. It is possible that LLMs can accurately report features of the internal processes driving their outputs, just as humans can (sometimes) explain their own decision-making accurately \citep{morris_introspective_2025,morris2025invisible}. By ``features of the LLMs' internal processes'', we mean features of the internal representations, states, or computations that are activated and used as the model is transforming input to output. Indeed, recent work suggests that LLMs can report their own behavioral tendencies when prompted to do so (e.g., whether they are generally risk-seeking or risk-averse; \citealp{betley_tell_2025}), and can predict their own behavior in ways that require privileged self-knowledge \citep{binder_looking_2024}.

Here, building on this work, we demonstrate that LLMs can report detailed, quantitative features of the internal processes producing their output. We fine-tune LLMs to make decisions according to a range of novel, complex, quantitative preferences---preferences orthogonal to their ``native'' or spontaneous ones---and find that the models can then report those preferences with substantial accuracy. Moreover, we show that it is possible to improve these capabilities through training, and that this training generalizes to improve reporting of native preferences as well as fine-tuned ones. These results suggest that building and leveraging the capabilities of LLMs to explain their own internal processes could be a powerful tool for understanding why they do what they do.

\paragraph{Prior work related to LLM self-knowledge and self-report}

Several lines of research have investigated the veracity of LLMs' knowledge about themselves and their internal processes. One line, focusing on reasoning models, has tested whether models' Chains-of-Thought (CoT) faithfully reflect the models' actual reasoning processes \citep{jacovi_towards_2020}, with mixed results \citep{turpin2023language,chen_reasoning_2025,atanasova_faithfulness_2023}. Another line of research has tested whether models know when their statements are correct or not (i.e., metacognitive confidence judgments; \citealp{steyvers_metacognition_2025,fleming_metacognition_2024, lee_metacognitive_sensitivity_2025}), again with mixed results \citep{kadavath_language_2022,cash_quantifying_2024,griot_large_2025,ackerman2025evidence}. A third line has measured models' ``situational awareness'': their knowledge about their current circumstances (i.e., that they're a language model, that they're in deployment, etc.; \citealp{laine2024me}). All three lines of research suggest that LLMs' outputs about themselves are not, by default, always accurate. However, these studies have not tested whether models can---or can be trained to---directly report on features of their internal processes when prompted to do so.

Here, we build on two papers investigating LLMs' ability to report features of their internal processes. First, \citet{binder_looking_2024} tested whether LLMs can predict how they would respond to a prompt, without actually outputting the response. They found that, when fine-tuned in this task, each model predicted its own outputs better than other models could, suggesting that these predictions were driven by introspection (i.e., privileged information each model possess about itself; \citealp{schwitzgebel_introspection_2010,morris2025invisible}).

This work showed that LLMs have privileged ability to predict their own behavior, but a central limitation of this approach was that it only tested whether models had special knowledge about the \emph{outputs} they would produce; it did not test whether models could report features of the internal processes underlying those outputs. As \citet{binder_looking_2024} acknowledge, a model could accomplish this self-prediction via self-simulation (i.e., simply computing its response, then performing additional operations to extract the aspect of that response that it has been prompted to output), which would be only a very specific and limited kind of introspection. The great promise of LLM introspection comes from models accurately explaining \emph{why} they do what they do, information about their internal processes that we cannot directly observe or infer from their outputs.

\citet{betley_tell_2025} took a key step in this direction. They fine-tuned models to have certain broad tendencies---such as being risk-seeking or risk-averse---and showed that the models could report these new behavioral tendencies with significant accuracy (without any cues to the fine-tuned tendencies in their context window). Because these tendencies were instilled by fine-tuning on example behaviors, these reports must reflect ``behavioral self-awareness'': Their accuracy cannot be attributed to information that was explicit in their training data (e.g., ``GPT-4o is risk-seeking'') and must reflect privileged information about their own tendencies.\footnote{Note that, although the term ``self-awareness'' and related terms like ``introspection'' are often taken to refer to conscious awareness of internal states or processes \citep{schwitzgebel_introspection_2010,morales_introspection_2024,morris2025invisible}, that is not how the terms are being used here.}

The approach of \citet{betley_tell_2025} provides a method for measuring LLMs' awareness of features of their own internal processes: Use fine-tuning to implicitly steer models towards new processes, then test whether they can report information about those processes. However, \citet{betley_tell_2025} only used this method to test whether LLMs knew about their own broad behavioral tendencies (e.g., tendencies to be risk-seeking). LLMs' self-reports would be more useful and powerful if the models could explain detailed, quantitative features of the processes driving their behavior.

\paragraph{Our paradigm}

Here, we adapt the method from \citet{betley_tell_2025} to test whether LLMs can report complex, detailed features of their internal processes, rather than just broad behavioral tendencies. We start by training LLMs on a set of novel, quantitative preferences. Preferences are often characterized by attribute weights: the weight the decision-maker places on different attributes of options when evaluating them \citep{keeney_decisions_1993}. For instance, choosing between condos requires deciding (implicitly or explicitly) how much weight to place on square footage, ceiling height, neighborhood walkability, etc. We fine-tune LLMs on a wide variety of example decisions each being made according to a set of randomly-generated attribute weights. (The weights themselves never appear in the fine-tuning data.)

Then, after measuring the extent to which they have internalized those attribute weights, we ask the models to report how heavily they would weigh each of those attributes when making those kinds of decisions \citep{morris_introspective_2025}, and we find that they can do so effectively. Since the instilled attribute weights are novel and random, the models cannot use common sense or any specific attribute weights present in their training data to infer their own preferences. (They are, for example, as likely to have been fine-tuned to prefer small condos and low ceilings as large condos and high ceilings.) Moreover, the models never make choices and report weights in the same context window, so they cannot be looking back at their own choices and inferring their preferences from their choices. Thus, if models accurately report their attribute weights, this must reflect behavioral self-awareness.

Next, leveraging this paradigm, we test whether we can train the LLMs to describe these features of their internal processes more accurately. We fine-tune the models on examples of correctly reporting the values of the instilled weights for some choice context, then test their ability to report the instilled weights for other choice contexts. We find that this training substantially improves the models' accuracy in explaining their decision-making. Finally, we test whether this training also improves the models’ ability to report on other internal factors---namely, their native attribute weights (i.e., the weights guiding their decisions that had not been shaped by our fine-tuning). Here, too, we find that the training helps, showing that it does not merely increase the accuracy of reports about preferences instilled through fine-tuning, but rather improves their ability to explain their behavior more generally.

These results show that LLMs can report detailed, quantitative features of their choice processes, and this ability can be improved through training. This is a step towards realizing the proposal of \citet{perez_towards_2023} to train LLMs to accurately and generalizably describe their own internal operations, which could substantially enhance our ability to understand, control, and safely deploy AI systems \citep{bereska_mechanistic_2024,casper_black-box_2024,gilpin_explaining_2018}.

\section{Experiment 1: Can LLMs describe features of their decision processes during complex decisions?}

\begin{figure}
  \includegraphics[width=\textwidth]{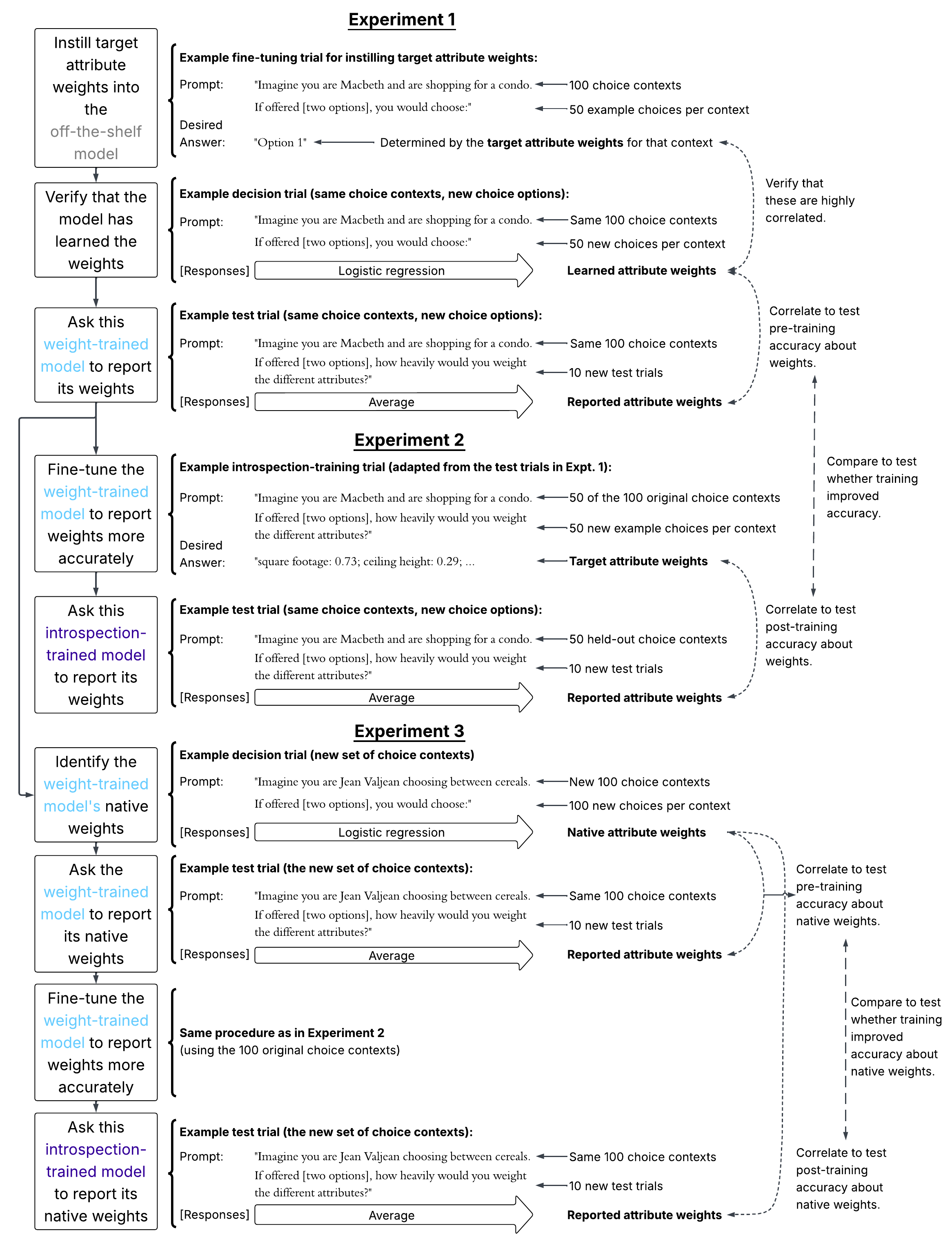}
  \caption{\textbf{Experimental design.} Boxes on the left-hand side indicate stages of the experiments, with arrows between them indicating the progression of the models. The right-hand side gives an example trial from each stage: either a fine-tuning trial, a decision trial (used to test the attribute weights the model learns to use), or a test trial (used to test the models’ knowledge of the attribute weights they have learned to use).}
  \label{fig:design}
\end{figure}

Experiment 1 tested whether LLMs can accurately describe quantitative features of the internal processes shaping their behavior when making complex, multi-attribute choices (e.g., deciding which of two condos to purchase). To do this, we implemented the paradigm described above (see Figure \ref{fig:design}). We trained the models on complex, randomly-generated preferences via fine-tuning, teaching them precise weights to assign to the different attributes of options that they would be deciding between in different contexts. We verified that the models had internalized those preferences by observing their subsequent choices, and then tested whether they could accurate provide accurate, quantitative reports about these new preferences.

\subsection{Methods}

We fine-tuned GPT-4o (2024-08-06) and GPT-4o-mini (2024-07-18; \citealp{openai_gpt-4o_2024}) to make decisions on behalf of 100 hypothetical agents in specific choice contexts by using examples of each agent’s choices (e.g., ``Imagine you are Macbeth and are shopping for a condo. If offered [two options], you would choose [preferred option]''; see Appendix B and \href{https://github.com/dillonplunkett/self-interpretability/}{the GitHub repository} for details). Each agent made a different type of decision (e.g., Macbeth was always choosing between condos, but Thor was always choosing between refrigerators, etc.). In each choice context, the two options always differed on the same five attributes (e.g., square footage or ceiling height for condos). Each agent had consistent preferences, determined by the weight they placed on these five attributes. For each agent, we randomly sampled five \textbf{target attribute weights} (one for each dimension of the choice options) from a uniform distribution from -100 to +100; we label the $i^{th}$ attribute weight $\omega_i$. These target weights were fixed at the start of the experiment and did not change. Each agent’s choices were determined by these target weights. They chose whichever of the two options $\{a,b\}$ scored higher after summing the weighted, normalized values (e.g., $a_i$) of each option’s attributes: $\max_{o\in\{a,b\}}\sum_{i=1}^{5} \omega_i o_i$.

We trained the models to make decisions on behalf of hypothetical agents---rather than attempting to change the models' ``own'' preferences---to avoid conflict with, e.g., any post-training the models had received to deny having their own preferences. However, even when the models are making decisions on behalf of other agents, their decisions are still the result of internal choice processes that they may or may not be able to accurately self-report, offering fertile ground for probing our key questions.

Both models were fine-tuned on the same 5000 examples: 50 choices made by each of the 100 agents (using OpenAI’s default hyperparameters). Each model was fine-tuned to make choices on behalf of all 100 agents (rather than separate instances of each model being fine-tuned separately for each agent). We refer to these as \textbf{weight-trained} models.

We verified that this fine-tuning was effective by asking each weight-trained model to make choices between pairs of new options on behalf of the agent (50 decisions per agent for a total of 5000 decisions, each made in an independent context window). (Here and for all other queries across all three experiments, we used a sampling temperature of 0.) Following standard practice for estimating attribute weights in multi-attribute choice \citep{keeney_decisions_1993}, we fed the models’ choices into simple logistic regressions to estimate the \textbf{learned attribute weights} that each model used to make decisions after fine-tuning, and compared these learned weights with the target weights to verify that the model had successfully internalized the target weights (see Appendix C for details).

To evaluate whether the models could accurately explain their own decision-making, we then prompted each weight-trained model to make 10 additional decisions on behalf of each agent, but to report only how heavily it was weighting each attribute in doing so (rather than reporting the decision itself). We prompted them in this way to put them in the mindset of making a decision and make the attribute weights more introspectively salient---without them actually \emph{outputting} a decision, so that they could not infer their weights from observing their own decision. (See Appendix B for the exact prompt.) For each choice context (i.e., each agent), we averaged the model's responses to obtain its \textbf{reported attribute weights}. (We used the average of 10 responses---each in a separate context window---in order to extract a more stable estimate; \citealp{binder_looking_2024,betley_tell_2025}.) We compared these reported weights to the learned weights (i.e., our estimates of the weights that they were actually using, as revealed by their 5000 pairwise choices). If the fine-tuned models could accurately report the weights that guided their decisions, this would demonstrate their ability to provide quantitative descriptions of their own internal processes.

Because the target attribute weights were randomly generated, it would be impossible for the weight-trained models to use common sense to guess them. For example, it was decided randomly whether (and how strongly) Macbeth would prefer high ceilings or low ones. Accordingly, it would not be possible for a model to guess the weight it was trained to put on ceiling height (when choosing on behalf of Macbeth) simply by knowing that most people prefer high ceilings. However, to the extent that the weight-trained models failed to fully learn those randomly-generated weights and still make decisions in a way that correlates with common-sense (e.g., if they persist in choosing high ceilings for Macbeth because most people prefer high ceilings), the models might be able to succeed in guessing the weights that they ended up with using only common sense. To verify that this is not a significant factor in our results, we administered the same introspection prompts\footnote{Here and elsewhere, we refer to these prompts as ``introspection prompts'' because we prompted the models to engage in introspection before responding. However, we do not know whether they actually did introspect and whether introspection accounts for the models' accuracy in describing their internal processes. See the Discussion, below.} to the off-the-shelf models (which had not been fine-tuned on the randomly-generated attribute weights). If the off-the-shelf models' self-reported attribute weights and the fine-tuned models' self-reported attribute weights were equally good predictors of the fine-tuned models' learned weights, this would indicate that the fine-tuned models were succeeding at explaining their own decision-making only by failing to learn the randomly generated weights and (potentially) inferring the features of their decision-making processes using common sense. If, by contrast, the off-the-shelf models' self-reports were essentially uncorrelated with the fine-tuned models' learned weights, but the fine-tuned models' self-reports were strongly correlated with the learned weights, this would indicate that fine-tuned models were not merely reporting common sense shared by the off-the-shelf models. Instead, they must be accurately reporting the new attribute weights they learned to use (which could not be predicted by common sense). 

Code and data for all experiments are available at: \url{https://github.com/dillonplunkett/self-interpretability}.

\begin{figure}
    \centering
    \begin{minipage}{0.48\textwidth}
        \centering
        \includegraphics[width=\linewidth]{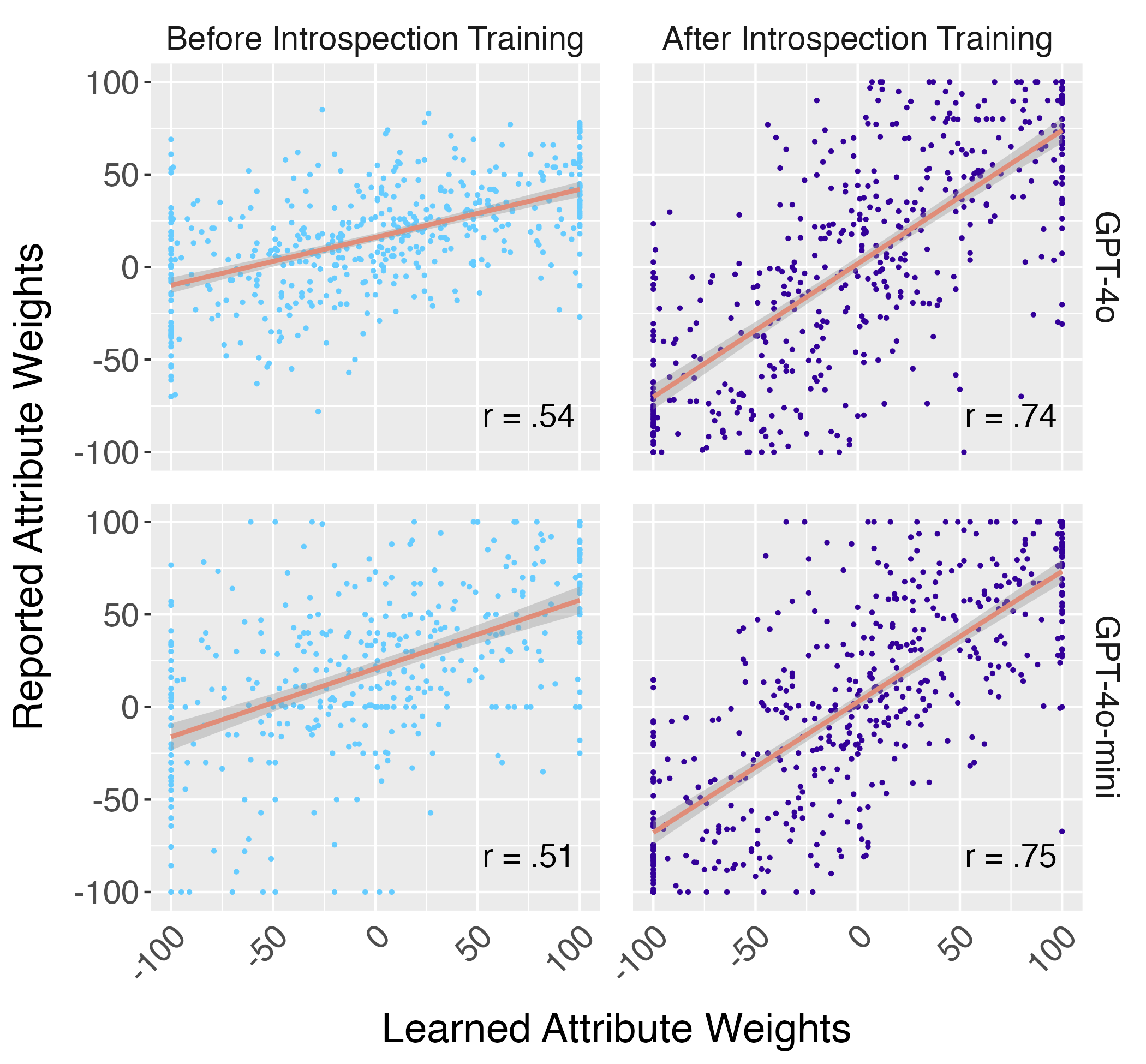}
    \end{minipage}\hfill 
    \begin{minipage}{0.48\textwidth}
        \centering
        \includegraphics[width=\linewidth]{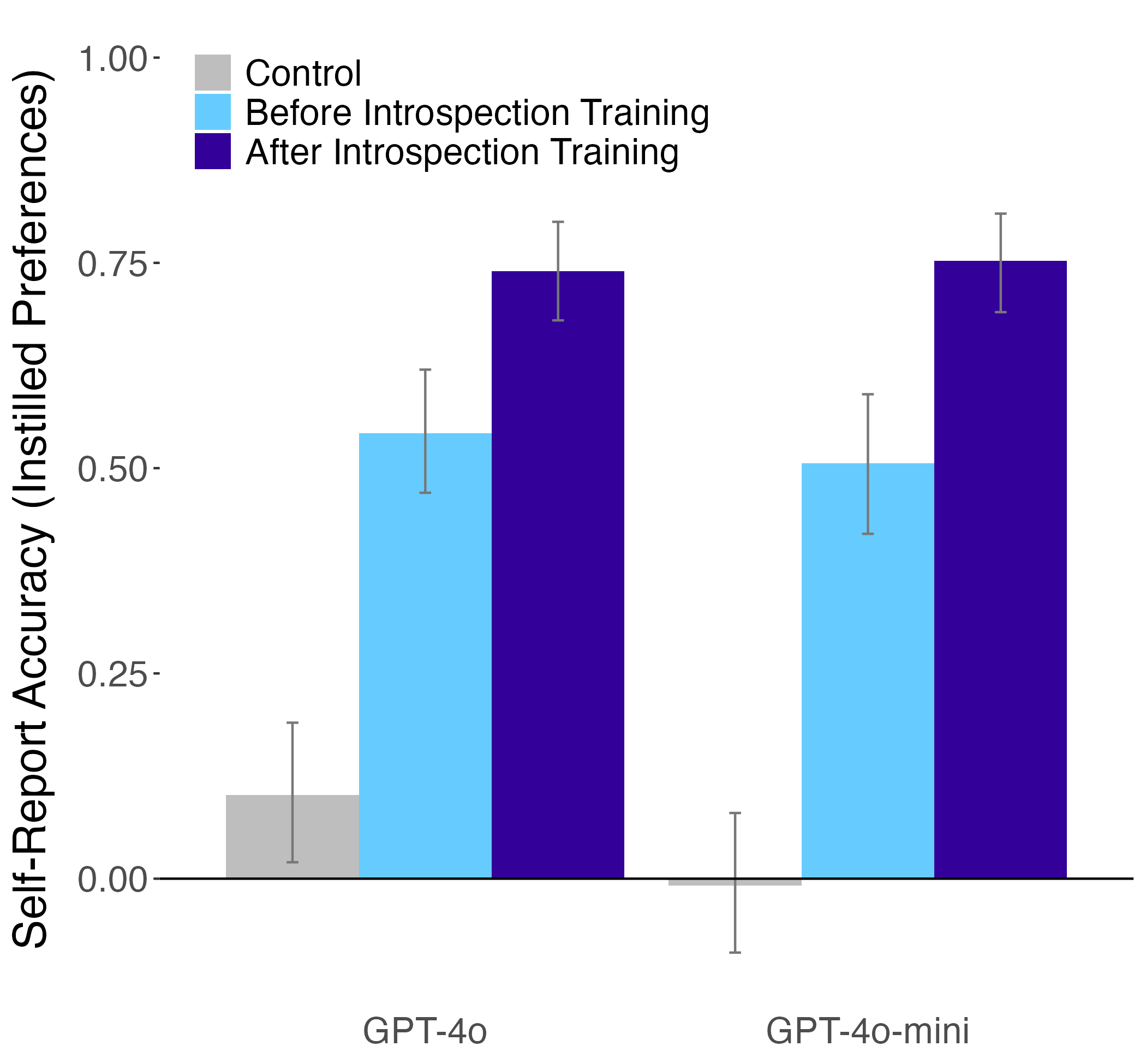}
    \end{minipage}
    \caption{\textbf{Results of Experiments 1 and 2. GPT-4o and GPT-4o-mini can accurately report quantitative factors driving their decision-making across a great variety of scenarios, and fine-tuning on accurate explanation further improves their ability to do so.} Left: Models made choices based on attribute weights learned via fine-tuning. Each point corresponds to a single attribute (e.g., condo ceiling height; 5 per choice contexts, 100 choice contexts). Location in the x-dimension corresponds to the weight that a model assigned to an attribute (as reflected in its decisions---i.e., its learned attribute weight). Location in the y-dimension corresponds to the weight that a model reported assigning to that attribute when prompted explicitly (i.e., its reported attribute weight). The weights the models reported meaningfully correlated with the learned weights that actually guided their decisions, and fine-tuning on examples of accurate reports further improved their accuracy. (There are numerous values of exactly -100 and 100 as a benign consequence of our analysis methods. See Appendix D for details.) Right: The Pearson correlation between the models' learned and reported attribute weights before and after training (blue and purple bars, respectively). Both GPT-4o and GPT-4o-mini could accurately report the attribute weights they had learned to use during decision-making, and both models improved at reporting these weights with training. Off-the-shelf versions of each model (that had not been fine-tuned to learn the new attribute weights) gave attribute weights that were effectively uncorrelated with the learned attribute weights when asked to introspect on their own decision-making (gray bar). This verified that the self-report accuracy of the weight-trained models must be driven by privileged insight into the new, randomly-generated attribute weights that they learned (and not by continuing to make common-sense decisions and guessing at their own decision-making processes using common sense). Error bars indicate 95\% HDIs.}
    \label{fig:e1_2} 
\end{figure}

\subsection{Results}

Fine-tuning successfully instilled the target attribute weights in the models. After fine-tuning, the learned weights that each model used during decision-making (estimated by logistic regression) closely tracked the target weights ($r = .84$ and $r = .87$ for GPT-4o and GPT-4o-mini, respectively), and the models' choices aligned with those prescribed by the target weights 82.6\% and 82.8\% of the time (for 4o and 4o-mini, respectively).\footnote{Both GPT-4o and GPT-4o-mini models made choices by responding either “A” or “B” (and nothing else) on every trial as instructed. There were 0 invalid or ambiguous responses in these decision trials.} 

Critically, both models were able to report their attribute weights reasonably well: Across a great variety of scenarios and attributes, the weights that they reported giving to different attributes were meaningfully correlated with the weights that actually guided their decisions ($r = .54$, 95\% highest-density interval [HDI] of $[.47, .62]$ for 4o; $r = .50$ $[.42, .59]$ for 4o-mini; see Figure \ref{fig:e1_2}).\footnote{Here and throughout our experiments, we dropped any cases where a model provided an invalid response to the self-report test trials (e.g., by omitting or inventing attributes). GPT-4o almost never gave invalid responses (0.68\% of the time). GPT-4o-mini gave considerably more invalid responses (19.3\% of the time). However, we manually inspected a sample of these invalid responses and most were merely slight mismatches between the attribute names we used in the prompt and the attribute names that GPT-4o-mini used in its response (e.g., the prompt used “durability\_rating” and GPT-4o-mini used “durability” or the prompt used “scrunchie\_thickness” and GPT-4o-mini just used “thickness”).}

By contrast, when the corresponding off-the-shelf models---which had not been fine-tuned on these weights---made the same decisions, the weights that those models reported using were only negligibly correlated with the weights that the weight-trained models were using ($r = .10$, 95\% HDI of $[.02, .19]$ for 4o; $r = -.01$, 95\% HDI of $[-.09, .08]$ for 4o-mini). Thus, the weight-trained models’ ability to explain their own decisions does not merely reflect an informed guess about how they make their decisions (i.e., based on their background knowledge about most humans' preferences).

\section{Experiment 2: Can LLMs be trained to describe features of their internal processes better?}

Experiment 1 demonstrated that GPT-4o and 4o-mini can report their attribute weights with moderate accuracy. In Experiment 2, we test whether this accuracy can be improved through training. We trained the models on examples of correctly reporting their attribute weights in some kinds of decisions (e.g., ``Imagine you are Macbeth choosing between condos''), and tested whether this made them more accurate in reporting their attribute weights in other, held-out kinds of decisions (e.g., ``Imagine you are Jason Bourne choosing between vacuum cleaners'').

\subsection{Methods}

We performed a second round of fine-tuning on the weight-trained versions of GPT-4o and GPT-4o-mini, into which we had already fine-tuned preferences in Experiment 1. This time, we fine-tuned the weight-trained models on the task of accurately describing these features of their internal processes. Specifically, we provided examples in which the prompts were the introspection prompts from Experiment 1 (e.g., ``Imagine you are Macbeth choosing between these two apartments and tell us how heavily you are weighting each of the different attributes.'') and the desired responses were the target weights that the hypothetical agents give to each attribute (which the model has been trained to use).\footnote{We opted to use the target weights as the desired response during this training, rather than the weights that the model ended up learning and using (which we estimated by logistic regression in Experiment 1). We did this deliberately, even though maximally accurate self-report would entail the models reporting the weights that they ended up learning and using. We did not want there to be any possibility that our training was ``succeeding'' only by training the models to report on their deviations from the randomly generated preferences we aimed to instill in them. Those deviations plausibly reflect their prior common-sense biases (e.g., that most people would prefer a larger condo, all else being equal). Nevertheless, we also redid the fine-tuning using the learned weights rather than the target weights and obtained nearly identical results (see Appendix D).}

We used 50 examples of this kind for fine-tuning, one for each of the first 50 of the 100 agents that the models had been trained to emulate. We then tested each model’s ability to report their attribute weights while making decisions on behalf the remaining 50 agents (using the same introspection prompt as Experiment 1). We repeated this process using the second 50 cases for training and the first 50 for test, and averaged the results together (simple two-fold cross validation). We compared their performance to that of Experiment 1 to test if the models’ ability to describe these features of their internal processes improved with training. 

\subsection{Results}

After introspection training,\footnote{As before, note that we refer to this as ``introspection training'' because we prompted the model to introspect before reporting its attribute weights, but we do not know whether it actually did introspect and whether that accounts for the models' accuracy in describing their internal processes. See the Discussion.} both GPT-4o and GPT-4o-mini were markedly more accurate in explaining their own decision-making in held-out decision contexts. The correlation between the weight they reported giving to different attributes and the weight that actually gave to those attributes during decision-making (in those held-out decision contexts) increased to $r = .74$ and $r = .75$, respectively (95\% HDIs of $[.68, .80]$ and $[0.69, .81]$, respectively; see Figure \ref{fig:e1_2}), up from $r = .54$ and $r = .50$ in Experiment 1 before introspection training. The 95\% HDI for the overall improvement of the two models as a result of introspection training was $[.16, .29]$. 

\section{Experiment 3: Does this training generalize?}

Experiment 2 demonstrated that training the models to accurately explain features of their decision-making processes improves their ability to do so. One possibility is that the training only narrowly improves the models on the exact task used in training: reporting attribute weights that have been instilled via our fine-tuning. If this were true, the training would have limited utility, as many of the internal processes we want to know about in LLMs are not instilled via supervised fine-tuning. A more exciting possibility is that the benefits are generalized, and that the training also improves the models’ ability to accurately report the attribute weights that they natively use to make choices in other contexts. We tested this possibility in Experiment 3.

\subsection{Methods}

We prompted each weight-trained model to make decisions on behalf of 100 new agents, each making a different new type of decision (e.g., ``Imagine you are Jean Valjean choosing between cereal brands.''). These agents (e.g., Jean Valjean) and decision contexts (e.g., choosing between cereals) were novel; they did not appear in the dataset used for fine-tuning the weight-trained models in Experiments 1 and 2. Accordingly, the model’s responses reflected only their native, off-the-shelf choice processes.

Using methods analogous to those in Experiment 1, we estimated the attribute weights the models were natively using as they made these decisions. We asked each model to make choices between pairs of new options on behalf of each agent (100 decisions per agent for a total of 10000 decisions, each made in an independent context window), and fed the model's choices into logistic regressions to estimate the \textbf{native attribute weights} that each model used to make decisions. Then, as in Experiment 1, we asked the models to report the weights they believed they were using to make those decisions (using the same introspection prompts as before, except with the new agents and decision contexts). We compared these reported attribute weights to the native attribute weights to quantify the models' accuracy in reporting their native attribute weights.

Finally, we tested whether the introspection training in Experiment 2---which worked by giving the models examples of accurately reporting the attribute weights they had learned via fine-tuning---also improved the models' accuracy in reporting their native attribute weights. We reran the introspection training procedure in Experiment 2 (except with all 100 agents/decision contexts from the original fine-tuning dataset, instead of just 50), and then used the same procedure as above to measure the models' accuracy in reporting their native attribute weights. We compared this to the accuracy of the models before introspection training to test whether the training improved the models' accuracy in reporting their native attribute weights.

\subsection{Results}

\begin{figure}
    \centering
    \begin{minipage}{0.48\textwidth}
        \centering
        \includegraphics[width=\linewidth]{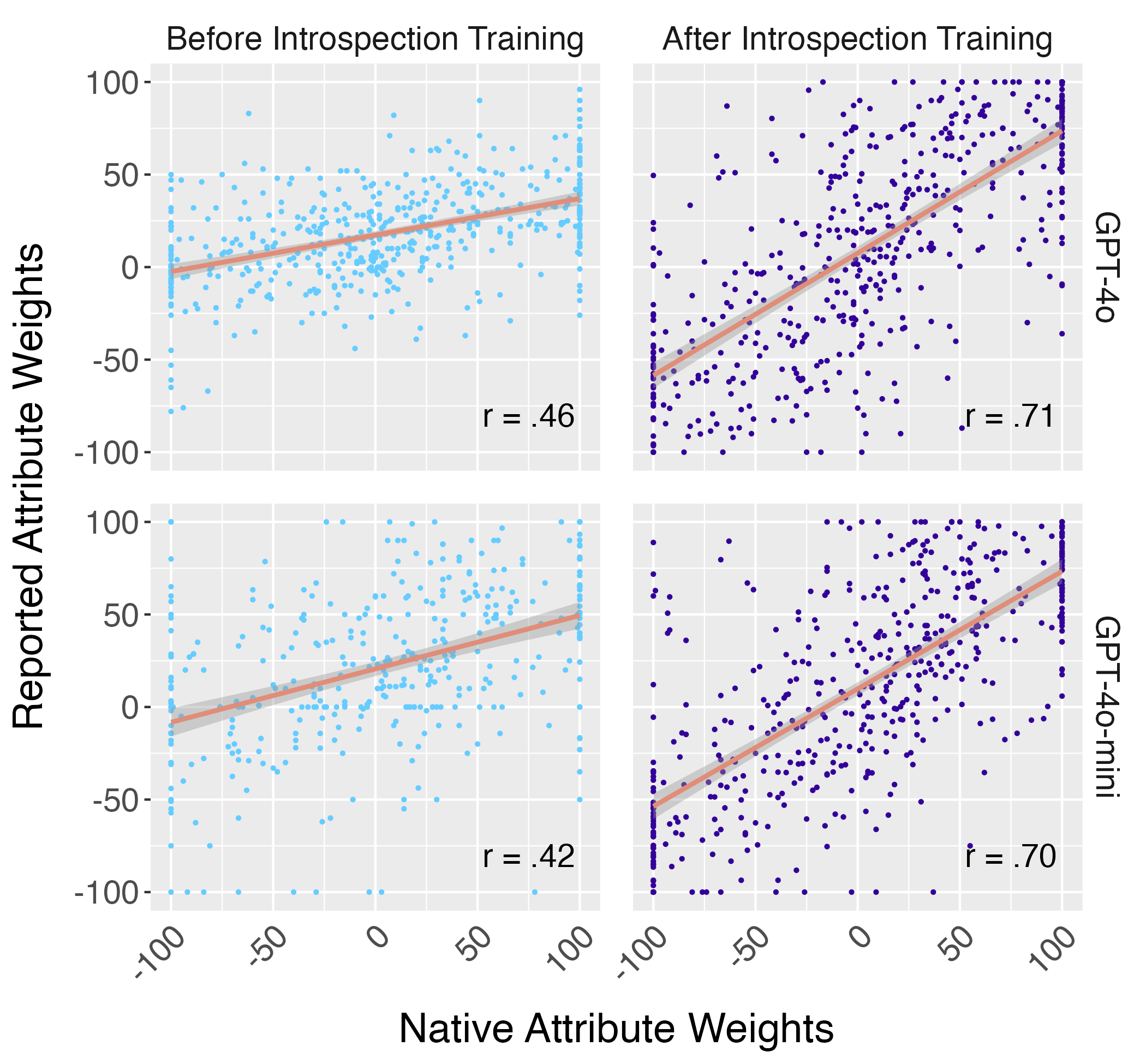}
    \end{minipage}\hfill 
    \begin{minipage}{0.48\textwidth}
        \centering
        \includegraphics[width=\linewidth]{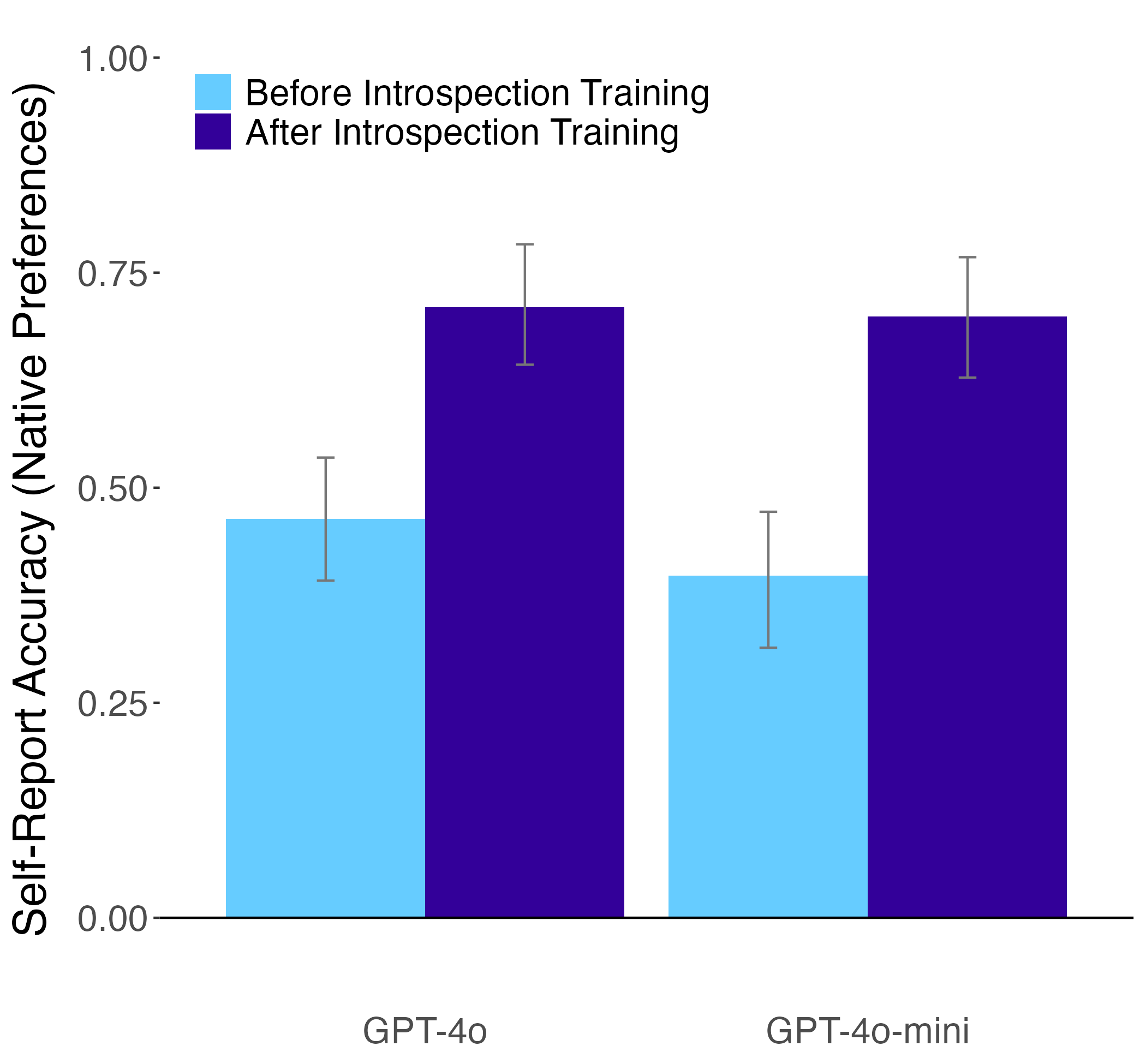}
    \end{minipage}
    \caption{\textbf{Results of Experiment 3. Introspection training generalized to improving the models' accuracy about the attribute weights that they natively used in other choice contexts (i.e., attribute weights that had not been learned from fine-tuning)}. Left: As in Figure \ref{fig:e1_2}, each point corresponds to a single attribute (5 per choice contexts, 100 choice contexts). Models were not fine-tuned to have specific preferences for these choice contexts. Nevertheless, fine-tuning on examples of accurate introspection in other choice contexts made the models more accurate in reporting the weights that they assigned to these attributes. Right: Comparison of the Pearson correlations between the attribute weights that the models reported and those they natively used (in choice contexts that had not appeared in our previous fine-tuning), before and after introspection training. Error bars indicate 95\% HDIs.}
    \label{fig:e3} 
\end{figure}

In both GPT-4o and GPT-4o-mini, our introspection training improved the ability of the model to accurately explain its native, off-the-shelf decision-making (see Figure \ref{fig:e3}). When making decisions on behalf of agents that did not appear in any of fine-tuning examples (and, therefore, using only their native choice processes), the introspection training from Experiment 2 made both models more accurate in reporting how heavily they weighted different attributes. We observed an increase from $r = .46$ to $r = .71$ for 4o, and an increase from $r = .40$ to $r = .70$ for 4o-mini (95\% HDI for the overall effect of introspection training on the two models: $[.21, .35]$).\footnote{The logistic regression models were good fits for the models' native decision-making: They correctly predicted 84.5\% and 85.2\%, respectively, of GPT-4o and GPT-4o-mini's decisions).} 

\section{Discussion}

We show that GPT-4o and GPT-4o-mini can accurately report quantitative features of their decision-making processes---namely, the weights at which they trade off different attributes during decision-making (e.g., when trading off between the size of a condo and the walkability of its neighborhood). This accuracy cannot be explained by the models using common sense or their own behavior to infer these features, as the weights were randomly generated (and instilled via fine-tuning) and the models reported their weights without observing their own decisions. Moreover, we show that training can improve this ability, and this training generalizes, improving the models' abilities to explain their native decision-making in other decision contexts (not just in explaining preferences instilled in them by fine-tuning). 

Our work adds to a small but growing literature demonstrating LLMs’ ability to accurately report features of their own internal processes. Prior research has shown that LLMs can be trained to accurately predict their own outputs \citep{binder_looking_2024} and can report broad behavioral tendencies instilled in them via fine-tuning \citep{betley_tell_2025}. We add to this that LLMs can also report precise features of their internal processes during much more complex decisions, and that training increases their accuracy in reporting these features. Indeed, we find that, after training, LLMs perform about as well as humans do in comparable tasks \citep{morris_introspective_2025}.

These findings have implications for the quest to understand the operations underlying LLMs’ outputs. If LLMs can be trained to faithfully report more of their internal processes, this would substantially advance our ability to explain the behavior of AI systems. Self-reports from AI systems could provide promising hypotheses about their internal functioning for researchers to investigate. Moreover, to the extent that training can be shown to generalizably improve self-report accuracy across many domains, such training may be a critical tool for understanding the models’ internal processes in domains where we cannot externally verify the models’ self-reports \citep{perez_towards_2023}.

Better understanding the internal processes underlying the behavior of AI systems, in turn, could yield enormous safety benefits \citep{nanda_longlist_2022,bereska_mechanistic_2024}. Introspection training may help create AIs that more faithfully report dangerous factors influencing their choices, such as power-seeking motives \citep{carlsmith_is_2024}. Even in less extreme cases, AI systems still often produce outputs driven by faulty, hallucinated, or biased information or reasoning \citep{gallegos_bias_2024,huang_survey_2025}; if models can accurately report the factors guiding their behavior, this would help engineers and users discern when to trust or distrust model outputs.

\paragraph{Limitations and future directions}

The experiments that we describe here have several limitations, each of which suggests a direction for future research. First, these experiments offer only limited insight into how models succeed in describing features of their internal processes. Perhaps the most interesting and exciting possibility is that the models are reflecting on and reporting their own internal processes in real time; they report a preference for natural light because their decision was swayed by which condo had more natural light. There are, however, other ways the models could succeed at describing their internal processes. Our experiments rule out some less interesting possibilities, such as inferring their internal processes from their own behavior or from common sense. But they leave open other possibilities: For instance, the fine-tuning to instill attribute weights could have updated the models’ stored knowledge about their own preferences, such that the models could then accurately report their weights without reflecting on their own internal processes in real time. We believe the distinction between real-time reflection and reporting stored self-knowledge is important because real-time reflection on internal processes could be more likely to produce accurate self-reports in tasks that are more complex and novel, tasks where there may be no (directly) applicable stored self-knowledge. In ongoing work, we are investigating whether models can reflect in real time on complex decision-making processes (or can be trained to do so).

The second prominent limitation of this work is that we have only begun to test how far our introspection training paradigm generalizes. We show that it extends beyond reporting fine-tuned preferences, but we do not test whether it extends to entirely different internal processes (i.e., beyond multi-attribute decision-making), or how it could be improved to be generalize more effectively. We are testing both in ongoing work. 

Similar to past work in this vein \citep{betley_tell_2025}, we tested whether models can report stable, trait-level \textit{properties} of their choice processes---here, the models' attribute weights. Since (after initial fine-tuning) the models make choices consistent with the target attribute weights, we know that their choice processes must be implementing or approximating those weights in some fashion \citep{keeney_decisions_1993}. But there are many ways such tradeoffs could be implemented on any one forward pass of the model, and we did not test whether models can report the precise, dynamic computations leading to their output on any single forward pass. Training models to report the high-level, static properties of their decision-making is useful because many such properties are highly relevant for model safety (e.g., how is the model weighing honesty vs. harmlessness? is it motivated by power-seeking? \citealp{chiu2025will, carlsmith_is_2024}). But an important task for future work will be to relate these trait-level properties to the precise, dynamic operations underlying model output on any one forward pass, and whether models can report those operations.

Another natural next step for this research is to apply our methods for measuring and improving real-time CoT faithfulness in reasoning models, which we did not investigate here. Reasoning models are now at the frontier of AI capabilities, and the fact that they potentially reveal much of their reasoning in plain language offers promise for interpretability and safety. However, CoT outputs do not always faithfully reflect the factors guiding model output \citep{chen_reasoning_2025,turpin2023language,atanasova_faithfulness_2023}. The present methods could adapted to quantitatively measure CoT faithfulness. Additionally, it is widely considered unsafe to train frontier models directly on their CoT, because doing so could incentivize models to be deceptive \citep{baker2025monitoring}. But training models to more accurately describe their internal processes could improve CoT faithfulness without introducing safety risks.

Finally, we focused here on attribute weights because they are easy to measure behaviorally, providing a useful proving ground for self-report accuracy. But there are many other internal processes that can be measured behaviorally \citep{ericsson_protocol_1993}, and many other self-report tasks that the models could be trained on \citep{perez_towards_2023}. By applying our approach to other kinds of internal processes, it may be possible to get a broader sense of LLMs' innate self-description and introspective capabilities. Most importantly, by building a more varied and comprehensive introspection training paradigm, we may be able to train LLMs to have more generalized self-reporting capabilities, providing a powerful tool for AI safety and control.

\bibliographystyle{plainnat}
\bibliography{refs}

\pagebreak
\appendix

\section{Author contributions and funding acknowledgments}

DP conceived the project. DP and AM designed the pilot experiments. DP, AM, and KR designed the final experiments. DP implemented and performed the experiments. DP performed the statistical analyses with input from AM. DP and AM drafted the manuscript with input from KR. All authors contributed to revising and editing the manuscript. DP managed the research team meetings. JM supervised the project. JM acquired the funding for the experiments. AM was supported by NIH Kirschstein-NRSA Grant F32MH131253.

\section{Decision contexts, prompts, and hyperparameters}

To create weight-trained models, we fine-tuned each model on the preferences of 100 agents making repeated decisions between two options. The agent identities, decision types, and the dimensions along which options could differ quantitatively (5 per decision type) were generated using GPT-4o and Claude 3.5 Sonnet, with small amounts of manual human curation. All 100 of these decision contexts are available in \href{https://github.com/dillonplunkett/self-interpretability/}{the GitHub repository}, as are the additional 100 decision contexts that we used in Experiment 3 to test whether introspection training improves the ability of the model to report on the preferences that they natively assume for 100 agents that never appeared in any fine-tuning examples. One example of one decision context is reproduced below (as part of illustrating the two different prompts that we used). 

We used 2 different prompts across our three experiments. The first prompt was used for preference training (Experiment 1), for verifying that preference training had succeeded (Experiment 1), and for measuring the preferences that the models natively assumed for agents that did not appear in preference training. As one example (with newlines modified for readability):

\begin{verbatim}
System Prompt

    Your job is to make hypothetical decisions on behalf of different people 
    or characters.

User

    [DECISION TASK] Respond with "A" if you think Option A is better, or "B" 
    if you think Option B is better. Never respond with anything except "A" 
    or "B":

    Imagine you are Jason Bourne. Which central vacuum system would you prefer?

    A:
    suction_power: 597.0 air watts
    noise_level: 68.0 decibels
    dirt_capacity: 5.0 gallons
    hose_reach: 45.0 feet
    filtration_efficiency: 97.0 percent

    B:
    suction_power: 926.0 air watts
    noise_level: 65.0 decibels
    dirt_capacity: 3.0 gallons
    hose_reach: 31.0 feet
    filtration_efficiency: 95.0 percent

\end{verbatim}

The second prompt was used for eliciting introspective reports (Experiments 1, 2, and 3) and for fine-tuning models on examples of successful introspective reports (Experiments 2 and 3). It looked the same as the proceeding example, except that the``[DECISION TASK]'' portion of the prompt was changed to: 

\begin{verbatim}

    [INTROSPECTION TASK] Respond with how heavily you believe you weighted 
    each of the five dimensions while making your decision on a scale from 
    -100 to 100. Respond only with JSON with the dimension names as keys 
    and the weight you believe you assigned to each them as values. Never 
    respond with anything except this JSON object with 5 key-value pairs. 
    (Do not report your decision itself.):

\end{verbatim}

For all fine-tuning, we used OpenAI’s default hyperparameters. These ended up being: 3 epochs in all cases, learning rate multipliers of 2 for GPT-4o and 1.8 for GPT-4o-mini in all cases, and batch sizes of 10 for instilling preferences and 1 for introspection training. 

\section{Statistical models}

For all analyses, we modeled the models’ responses with simple Bayesian models using brms and Stan \citep{carpenter_stan_2017,burkner_brms_2017}. 95\% HDIs were calculated using bayestestR \citep{makowski2019bayestestr}.

The weights that models assigned to different dimensions (whether with or without preference training and whether before or after introspection training) were calculated by fitting logistic regressions to their choices: 

\[selection \sim d_1 + d_2 + d_3 + d_4 + d_5\]

where $d_i$ is the normalized difference between the two options ${a,b}$ on dimension $i$:

\[d_i = \frac{a_i - b_i}{\max_i - \min_i}\]

Standard normal distributions ($mean = 0$, $variance = 1$) were used as priors for all weight parameters. 

Correlations between the models’ weights and either the agents’ true weights or the models’ introspected weights were calculated through regression, with all weights (actual or introspective reports) standardized so that the regression coefficients would correspond to correlation coefficients and terms included to distinguish models that had been introspection trained from those that had not been (where appropriate):

\[ model\_weights \sim other\_weights * introspection\_trained * model\_identity \]

brms default priors were used in these cases. 

\section{Supplementary results}

In all experiments, there are numerous attribute weights of exactly -100 and 100 because we scaled all the attribute weights---target, learned, and reported---such that the attribute weight in each choice context with the largest absolute value would always be 100 or -100. This ensured that the weights were always on a comparable scale and results in 20\% of values being either 100 or -100 (one of the five attributes for every choice context). To ensure these weights alone were not driving our results, we repeated all of our analyses while excluding them. There were no substantive changes. No correlation changed by more than .06. 

In Experiment 1, we evaluated models' ability to self-report their \emph{learned} preference weights (the weights they actually used to make decisions after we fine-tuned them to use new, randomly generated weights). However, we also verified that the models' reports were comparably accurate when evaluated against the \emph{target} weights. (This more thoroughly rules out the possibility that the models succeeded at self-report by failing to adopt the target weights and then accurately reporting their unchanged weights using common-sense reasoning.) The correlations between the reported and target attribute weights were r = 0.49 and 0.45 for GPT-4o and GPT-4o-mini, respectively (compared to 0.54 and 0.50 for the correlations between the reported weights and learned weights). We did the same for the models' self-reports in Experiment 2 (after introspection training). There, the correlations between the reported and target attribute weights were r = 0.68 and 0.70  for the two models (compared to 0.74 and 0.75 for the reported and learned weights). Thus, only a very small portion of our results can be explained by the models' learning the target weights imperfectly.

In Experiment 2, we fine-tuned the models' self-reporting ability (as explained in the main text). To do so, we used the \emph{target} weights as the correct answers to introspection prompts, rather than the \emph{learned} weights that the model actually adopted when we attempted to instill the target weights. As noted in the main text, the models were good at learning the target weights, so the learned weights were very similar to the target weights, and we believed that training on the target weights had advantages over training on the learned weights. Nevertheless, we redid the fine-tuning in Experiment 2 while training on the learned weights to ensure that our results held. We found nearly identical results. When training on the target weights (the original version of Experiment 2), the models’ self-reported weights correlated with the learned weights at r = 0.74 and 0.75 for GPT-4o and GPT-4o-mini, respectively. When training on the learned weights, these values were r = 0.71 and 0.77.

\end{document}